\documentclass[conference]{IEEEtran}
\IEEEoverridecommandlockouts
\usepackage{cite}
\usepackage{amsmath,amssymb,amsfonts}
\usepackage{algorithmic}
\usepackage{graphicx}
\usepackage{textcomp}
\usepackage{xcolor}
\usepackage{graphicx}
\usepackage{color}
\usepackage{subcaption}
\def\BibTeX{{\rm B\kern-.05em{\sc i\kern-.025em b}\kern-.08em
    T\kern-.1667em\lower.7ex\hbox{E}\kern-.125emX}}
\begin{document}

\title{
A Reconfigurable Rocker-Bogie Robot \\for High Step Climbing and Turning
}

\author{Kento Koizumi$^1$, Tomoaki Ohba$^2$, Yuta Saito$^3$, Takeya Morito$^4$ and Kenji Suzuki$^1$
\thanks{This work was supported by University of Tsukuba and Japan Society for the Promotion of Science (KAKENHI) under Grant 23H00485 and Grant 25KJ0657.}
\thanks{$^1$Kento Koizumi and Kenji Suzuki are with the Artificial Intelligence Laboratory, University of Tsukuba, Tsukuba, Japan (e-mail: {koizumi@ai.iit.tsukuba.ac.jp}; {kenji@ieee.org}).}
\thanks{$^2$Tomoaki Ohba is with the Ph.D. Program in Engineering Mechanics and Energy,
University of Tsukuba, Tsukuba, Japan (email: s2530198@u.tsukuba.ac.jp).}
\thanks{$^3$Yuta Saito is with the Beaver's Hive, 
Osaka, Japan (email: ys@beavers-hive.com).}
\thanks{$^4$Takeya Morito is with the Development \& Design Division, Team MarsMars, Osaka, Japan (email: xtremerobocon@gmail.com).}
}

\newcommand{\insfig}[4]{\begin{figure}[tb]
    \centering
    \includegraphics[width=#1]{#2}
    \caption{#3}
    \label{#4}
\end{figure}}

\maketitle

\begin{abstract}
This study proposes a reconfigurable rocker-bogie mechanism that achieves efficient turning motion with a small number of actuators while maintaining high step-climbing capability. By installing motors at the bogie joints and actively swinging up and down bogies, the system enables switching between four-wheel and six-wheel configurations. Omnidirectional wheels are mounted on the rear ends of the rockers, allowing smooth turning in the four-wheel configuration based on a differential-drive model. Experimental evaluation using a prototype robot demonstrated that the proposed mechanism achieves zero-radius turning at a speed more than five times that of a conventional rocker-bogie mechanism equipped with six non-steerable grip wheels, while requiring only approximately 17\% of the total average wheel torque. In addition, the robot successfully climbed a 40 cm step with an average climbing time of 6.4 s, confirming its high turning and step-climbing performance.
\end{abstract}

\begin{IEEEkeywords}
rocker-bogie mechanism, reconfigurable mechanism, step-climbing robot, zero-radius turning, omnidirectional wheel
\end{IEEEkeywords}

\section{Introduction}
Unmanned ground vehicles (UGVs) have been
deployed in various domains such as inspection, logistics, and hazardous environment operations\cite{He2023-nf}.
These environments often involve 
narrow pathways and uneven terrain, requiring UGVs to exhibit both high maneuverability in confined spaces and strong obstacle traversal capability\cite{Bruzzone2012}.

\insfig{\columnwidth}{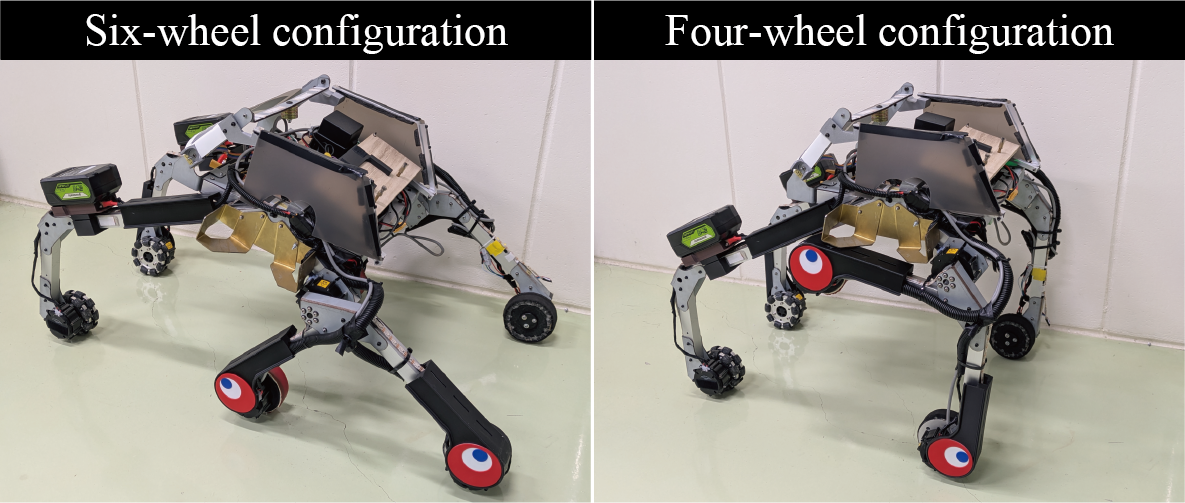}{Prototype robot with the proposed mechanism}{fig:md:prototype}

Numerous studies have investigated UGVs with high rough-terrain mobility\cite{He2023-nf, Bruzzone2012}. Among them, wheel-leg complex mobile-type UGVs have attracted considerable attention as robotic systems that achieve both high energy efficiency and superior obstacle traversal capability. In particular, the rocker-bogie mechanism
enables a balance between high rough-terrain mobility and turning performance\cite{The-Rover-Team1997-nz, Lindemann2005}. However, conventional rocker-bogie mechanisms typically require a large number of motors more than or equal to ten which leads to increased mechanical and control complexity as well as higher manufacturing costs. Consequently, various modified rocker-bogie mechanisms have been proposed to enhance functionality or simplify the system design\cite{Choi2012-mq, Lim2022-sx, Choe2017}.

Meanwhile, wheels without steering mechanisms are forced to slip laterally during turning, leading to significant torque loss and increased energy consumption \cite{Shamah1999, Visconte2021-cr}. As a result, conventional rocker-bogie robots without steering mechanisms experience substantial lateral slip, which increases torque demand and structural loads.
To address this issue, several rocker-bogie-based mechanisms have been proposed to improve both step-climbing and turning performance\cite{Chugo2005-vi, Takita2005-jh, Bruzzone2014}.
Siegwart et al. proposed a rhombus-shaped rocker-bogie configuration that demonstrated high step-climbing capability; however, noticeable wheel slip occurs during turning maneuvers\cite{Siegwart2002-vf}. 
Tokudome et al. proposed a rocker-bogie mechanism using omnidirectional wheels to reduce the number of motors; however, multiple omnidirectional wheels may increase slip during locomotion\cite{Tokudome2025-ub}.
Visconte et al. proposed a mechanism that alters the number of ground-contact wheels by lifting the bogie wheels; however, its step-climbing and turning performance were not reported\cite{Visconte2021-cr}.
These limitations suggest that a single mechanical configuration may not be sufficient to achieve consistently high mobility performance across different locomotion tasks.

In this study, we aim to develop a novel rocker-bogie mechanism that achieves both high step-climbing capability and high turning performance with a small number of motors. 
We focus on a reconfigurable mechanism that enables the robot to adaptively change its structure according to environmental demands.
To this end, we propose a modified rocker-bogie mechanism that combines omnidirectional and grip wheels with actuated bogie joints and introduces active bogie swing motion, enabling dynamic switching between locomotion modes.
We also fabricate a prototype robot based on the proposed mechanism, as shown in Fig.~\ref{fig:md:prototype}, and evaluate its step-climbing and turning performance through experimental validation.
The main contributions of this study are as follows. 
\begin{enumerate}
  \item We introduce a reconfigurable rocker-bogie mechanism that switches between four-wheel and six-wheel configurations to achieve both high step-climbing and turning performance with fewer actuators.
  \item We derive a mechanical model to estimate the torque required for the bogie swing-up motion and formulate the kinematics of the four-wheel configuration.
  \item We validated the proposed mechanism through performance experiments and the XROBOCON competition, demonstrating significantly improved turning efficiency while maintaining high step-climbing capability.
\end{enumerate}

\section{METHODOLOGY}
\subsection{Rocker-bogie mechanism with bogie swing-up motion}
\label{sec:rockerbogie}
The proposed rocker-bogie mechanism is shown in Fig.~\ref{fig:rb:cad}.
We attach grip wheels at both ends of each bogie and omnidirectional wheels at the rocker rear ends.
At the rocker front ends, we install motors as actuated joints connecting the rockers and the bogies. These motors are referred to as bogie joint motors throughout this paper.

\insfig{0.85\columnwidth}{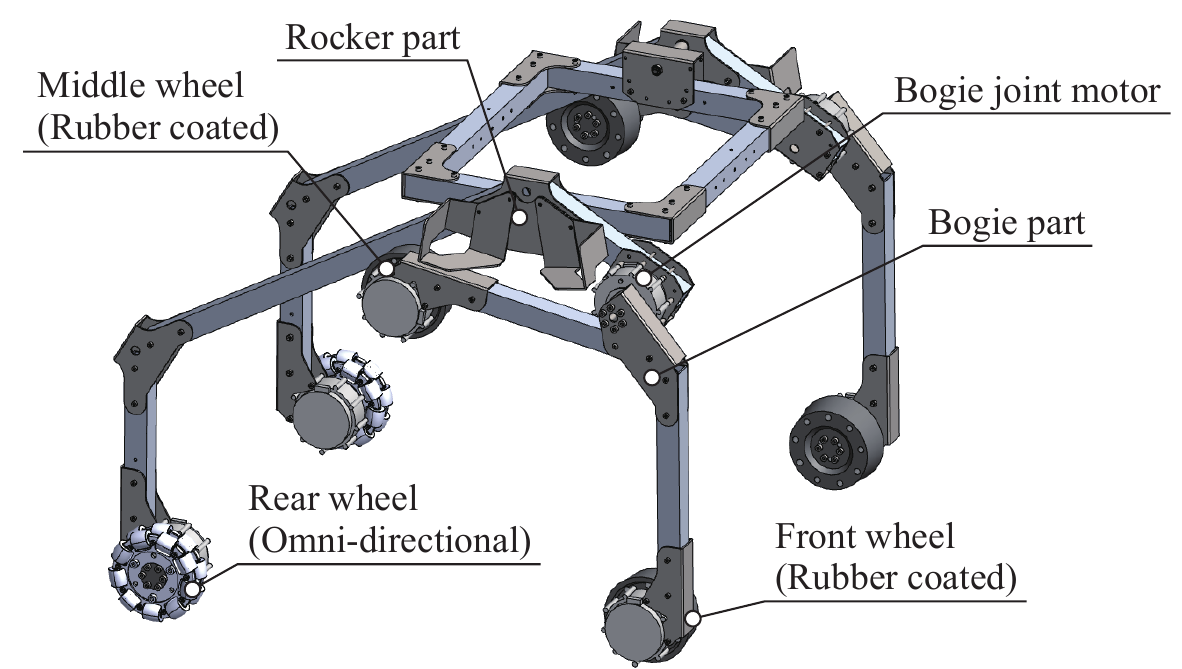}{CAD model of the proposed mechanism}{fig:rb:cad}

Owing to the bogie joint motors, the mechanism can switch between two configurations, as illustrated in Fig.~\ref{fig:md:prototype}. In the six-wheel configuration, the bogie joint motors are kept underactuated and all six wheels remain in contact with the ground.
This configuration behaves equivalently to a conventional rocker-bogie system and is capable of traversing steps.
During step climbing, the robot approaches the step from the bogie side, such that the rocker climbs onto the step last. This traversal direction is opposite to that of conventional rocker-bogie rovers.
In contrast, the four-wheel configuration actively rotates the bogies upward using the bogie joint motors, lifting the middle wheels off the ground.
Only the front and rear wheels are used for locomotion in this configuration.
Although this configuration sacrifices step-climbing capability, it enables smooth turning without a complex steering mechanism, owing to the omnidirectional wheels mounted at the rocker rear ends. 
Notably, the proposed mechanism requires only two additional actuators to achieve efficient turning, whereas conventional rovers typically require six actuators for the same function.
\subsection{Mechanical modeling for bogie swing-up motion}
\label{sub:Mechanical modeling for bogie swing-up motion}
\begin{figure}
    \centering
    \includegraphics[width=1.0\linewidth]{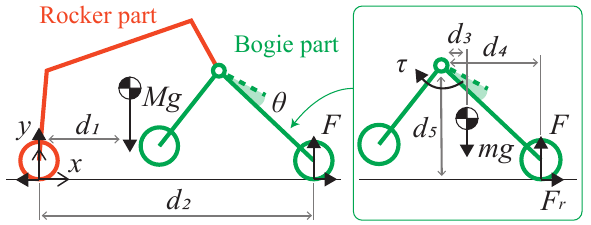}
    \caption{Mechanical model of the proposed mechanism}
    \label{fig:rover model}
\end{figure}

\begin{figure}
    \centering
    \includegraphics[width=.9\linewidth]{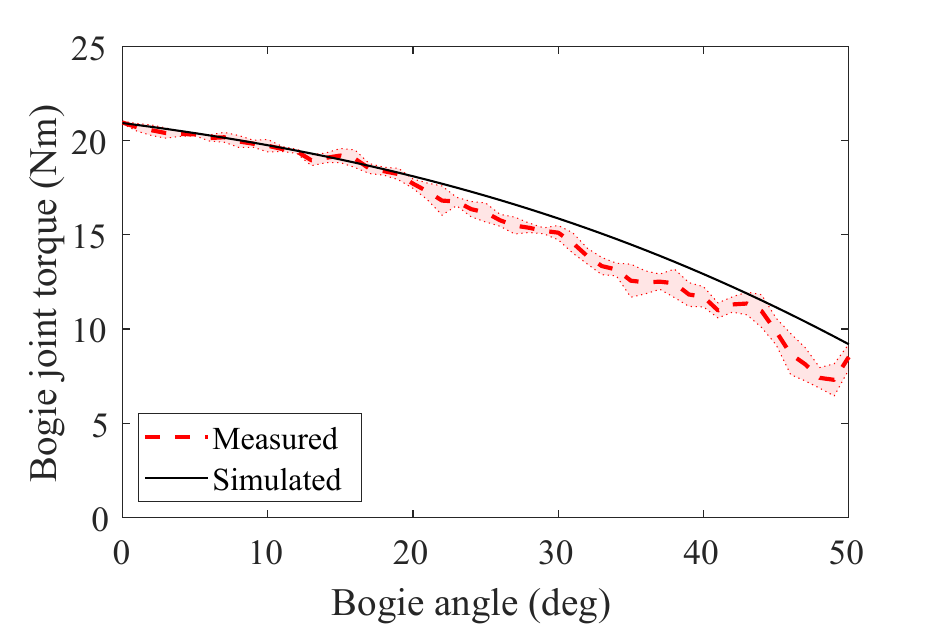}
    \caption{Relationship between the required bogie joint torque and the bogie angle during the bogie swing-up motion}
    \label{fig:bogie torque}
\end{figure}

To estimate the torque required for the bogie swing-up motion, we conduct a static analysis based on a simplified model of the mechanism shown in Fig.~\ref{fig:rover model}.
The figure represents the state in which the bogie is rotated by an angle $\theta$ from the initial six-wheel configuration, where all six wheels are in contact with the ground. The left side of the figure illustrates the forces acting on the entire robot, while the right side depicts the forces acting on the bogie. Let $M$ and $m$ denote the masses of the overall robot and bogie part, respectively. The horizontal distances from the rear-wheel axis to the overall center of gravity of the robot and to the front-wheel axis are defined as $d_1$ and $d_2$, respectively, and $g$ denotes the gravitational acceleration. On the left side of Fig.~\ref{fig:rover model}, the vertical reaction force at the front wheel $F$ can be expressed by (\ref{equ:front normal force}) based on the moment equilibrium on the rear-wheel axis.

\begin{equation}
\label{equ:front normal force}
    F = Mg\frac{d_1(\theta)}{d_2(\theta)}
\end{equation}

On the right side of Fig.~\ref{fig:rover model}, $d_3$ and $d_4$ denote the horizontal distances from the bogie joint to the bogie's center of gravity and to the front-wheel axis, respectively, and $d_5$ denotes the corresponding vertical distance to the front-wheel ground contact point.
A rolling resistance force $F_r$ acts at the front-wheel contact point and is modeled as $F_r=\mu_rF$, where $\mu_r$ is the rolling resistance coefficient. 
From the moment equilibrium on the bogie joint, the required torque $\tau$ on the bogie joint is derived by (\ref{equ:torque}). 
Note that when the point of application of the vertical force is located to the left of the bogie joint, the corresponding term in (\ref{equ:torque}) changes sign.

\begin{equation}
\label{equ:torque}
    \tau = -mgd_3(\theta)+F(\theta)d_4(\theta)+F_r(\theta)d_5(\theta)
\end{equation}

Based on (\ref{equ:torque}), we calculate the bogie joint torque as a function of the bogie angle through simulation, as shown in Fig.~\ref{fig:bogie torque}. 
The simulation incorporates the geometric parameters and weight of the prototype robot listed in Table.~\ref{tbl:md:dim}, with the rolling resistance coefficient $\mu_r$ set to 0.15. As shown in Fig.~\ref{fig:bogie torque}, the required bogie joint torque $\tau$ reaches its maximum when all six wheels are in contact with the ground and decreases as the bogie rotates upward. 
The dashed line in Fig.~\ref{fig:bogie torque} represents the torque estimated from experimental measurements using the prototype robot. The dashed line indicates the average of five measurements, while the red shaded region shows the standard deviation. 
Although the model does not account for factors such as joint friction and slip in the actual robot, the measured and simulated results show close agreement, supporting the validity of the proposed mechanical model.
These results indicate that the bogie swing-up motion can be achieved by selecting a motor capable of generating sufficient torque at the bogie joint in the initial six-wheel configuration.
Based on the simulation results, we determine the maximum required torque for bogie swing-up to be 21 Nm.

\subsection{Kinematics of the four-wheel configuration}
\label{sec:robot_design_ik}

\begin{figure}
    \centering
    \includegraphics[width=0.55\linewidth]{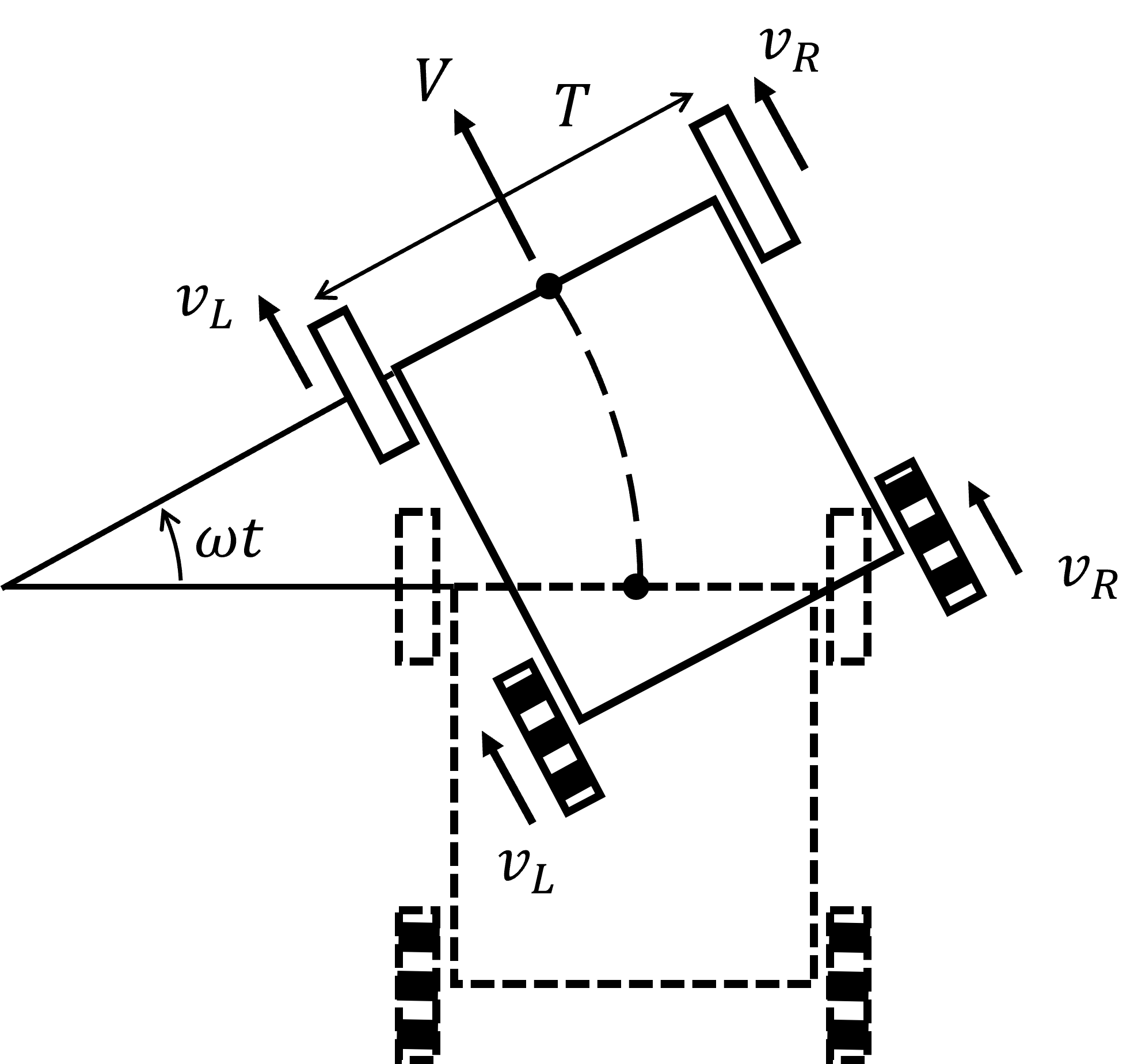}
    \vspace{-1mm}
    \caption{Turning movement of the four-wheel configuration}
    \label{fig:ik:movement}
\end{figure}

In the four-wheel configuration, we treat the two front wheels as a differential-drive pair, as illustrated in Fig.~\ref{fig:ik:movement}.
Moreover, we consider the two rear omnidirectional wheels to slip laterally while following the movement of the front wheels.
Therefore, given that right wheels share the same velocity $v_R$ and left wheels also share the same velocity $v_L$, 
the linear velocity $V$ and angular velocity $\omega$ at the midpoint between the front wheels are obtained from the standard differential-drive relations with front-wheel separation distance $T$, as shown in (\ref{eq:ik:vlvr}).
\begin{equation}
    \begin{gathered}
        V = \frac{v_R+v_L}{2} \qquad
        \omega = \frac{v_R-v_L}{T}
    \end{gathered}
    \label{eq:ik:vlvr}
\end{equation}

\section{PROTOTYPE ROBOT}
\subsection{Mechanical desigsn}
We design the robot to traverse steps up to 40 cm in height.
To avoid geometric interference and ensure 
continuous contact between the wheels and the ground or step surface,
we determine the frame lengths based on geometric constraints. In addition, we adjust the arrangement of onboard components so that the center of gravity is located near the middle wheels, 
improving stability and traction during four-wheel locomotion and step climbing.

\begin{table}[t]
    \vspace{2mm}
    \centering
    \caption{Dimensions and weight of the prototype robot}
    \vspace{-2mm}
    \begin{tabular}{c|c} \hline \hline
        Item & Value \\ \hline
        Width (cm) & 60 \\
        Length (cm) & 102 (six), 72 (four) \\
        Height (cm) & 55 (six), 64 (four) \\
        Wheelbase (cm) &  90 (six), 57 (four)\\
        Wheel diameter (cm) & 10 \\
        Total weight (kg) & 15.85 \\ \hline
    \end{tabular}
    \label{tbl:md:dim}
    \begin{minipage}{\columnwidth}
        \vspace{0.7em}
        \small
        \centering
        *(six) means six-wheel configuration and (four) means four-wheel configuration
    \end{minipage}
\end{table}

Fig.~\ref{fig:md:prototype} shows the prototype robot equipped with the proposed rocker-bogie mechanism. 
We use thermoplastic elastomers (TPE) as the rubber for the front wheels and commercially available table tennis rubber sheets for the middle wheels based on the results of comprehensive experiments.
We use the CyberGear motors (maximum torque: 12 Nm, from Xiaomi Corporation) for the bogie joints and six wheels, satisfying the required torque derived in Section \ref{sub:Mechanical modeling for bogie swing-up motion} and enabling the bogie swing-up motion. 
The robot is powered by two 24 V batteries mounted on the rear of the left and right rockers, each supplying power to the motors on the corresponding side.
To simplify the mechanism, we interconnect the left and right rockers through a single pivot joint rather than a differential mechanism.
The overall dimensions and weight of the prototype robot are listed in Table.~\ref{tbl:md:dim}.

\subsection{Control system}
Fig.~\ref{fig:cs:system} presents an overview of the robot's control system.
The robot is operated via a smartphone, which transmits motion commands to the onboard ESP32 microcontroller through Bluetooth Low Energy (BLE).
The ESP32 communicates with the CyberGears motors over the Controller Area Network (CAN), sending velocity or position commands corresponding to the received motion instructions.
Within each CyberGear, proportional-integral (PI) control loops are executed to achieve the commanded velocities or positions.
Sensor data including torque, velocity and position is returned from the CyberGears to the ESP32, and these measurements are then fowarded to a monitoring PC via WiFi for experimental evaluation.

\begin{figure}[t]
    \centering
    \begin{minipage}{0.45\columnwidth}
        \centering
        \includegraphics[width=\columnwidth]{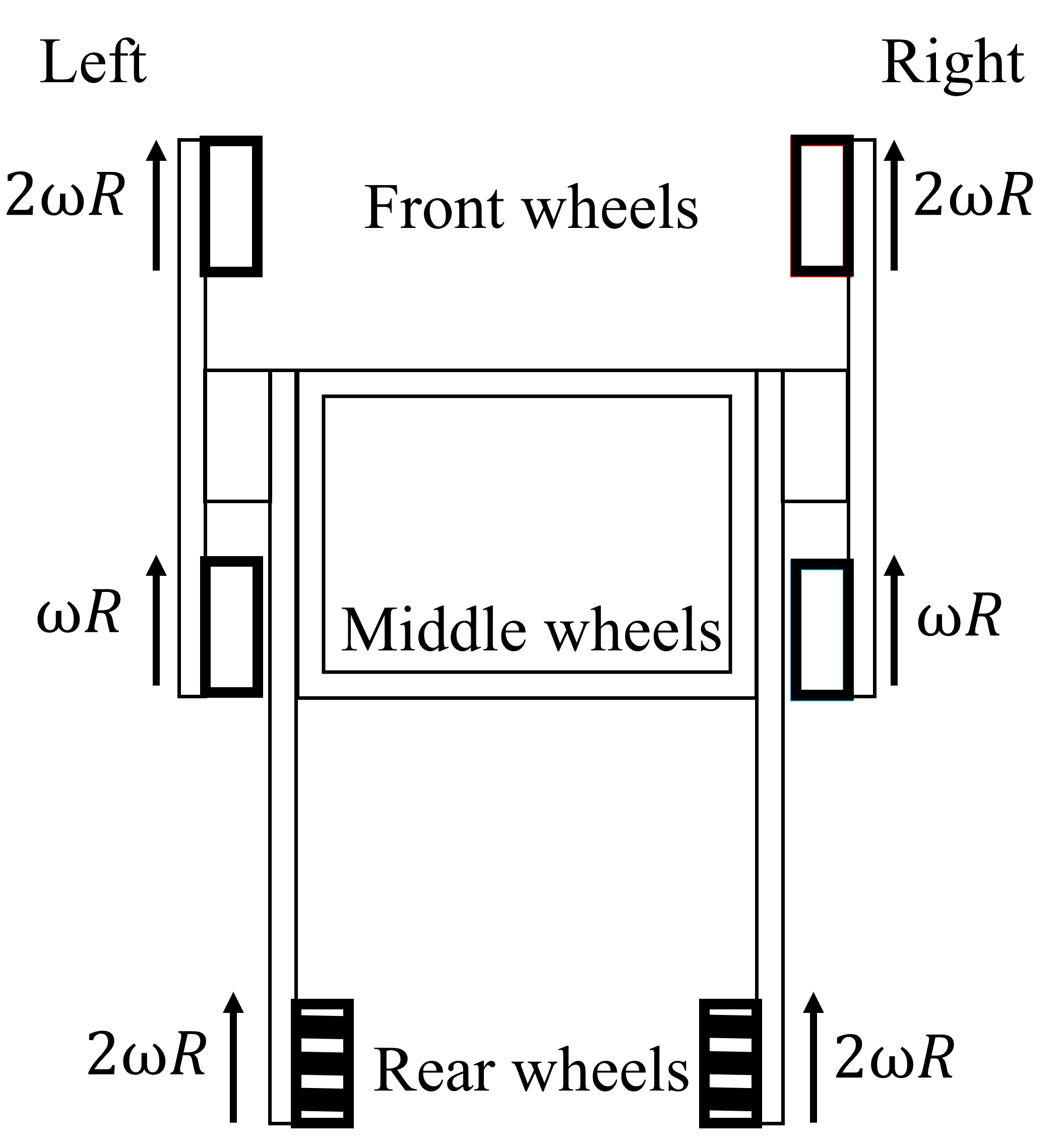}
        \subcaption{Go straight}
    \end{minipage}
    \begin{minipage}{0.45\columnwidth}
        \centering
        \includegraphics[width=\columnwidth]{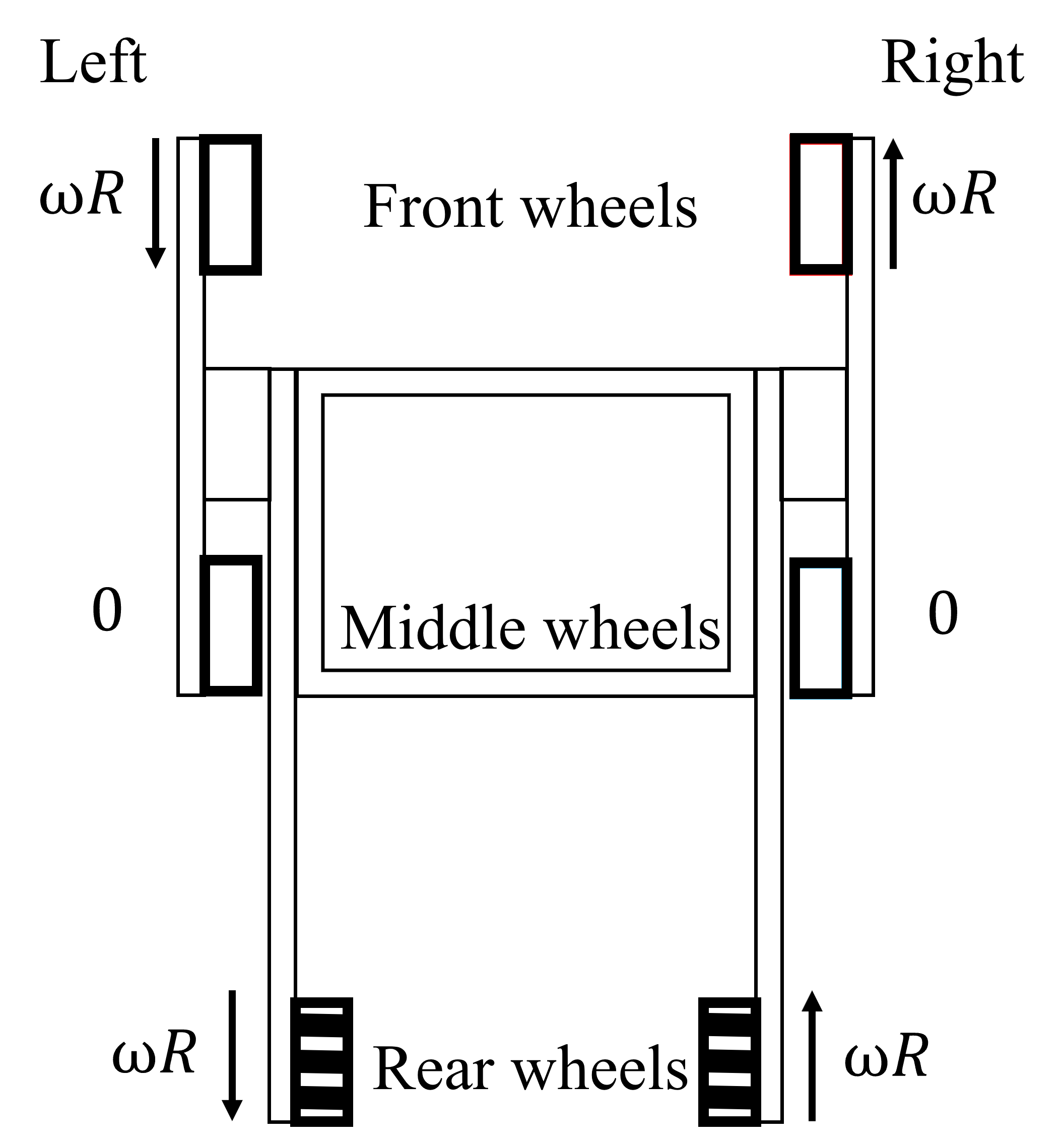}
        \subcaption{Turn at zero-radius}
    \end{minipage}
    \caption{Velocity assignments for each wheel when the robot goes straight and turns at zero-radius}
    \label{fig:cs:cmdsdr}
\end{figure}

Fig.~\ref{fig:cs:cmdsdr} illustrates the rotational velocity assignments for the CyberGears connected to the wheels when motion commands for going straight and turning at zero-radius are issued.
In Fig.~\ref{fig:cs:cmdsdr}, $\omega$ and $R$ denote the rotational velocity and the radius of the wheels respectively.
For the ``go straight'' command, the front and rear wheels are driven twice the velocity of the middle wheels $\omega$ to enhance the step-climbing performance.
For the ``turn at zero-radius'' command, the left and right wheels at both front and rear are driven at the opposite velocities, while the middle wheels are kept stationary.
This design reflects the fact that zero-radius turning is intended to be performed in the four-wheel configuration, where the middle wheels are not used for locomotion.

\begin{figure}[t]
    \centering
    \begin{minipage}{0.95\columnwidth}
        \centering
        \includegraphics[width=\columnwidth]{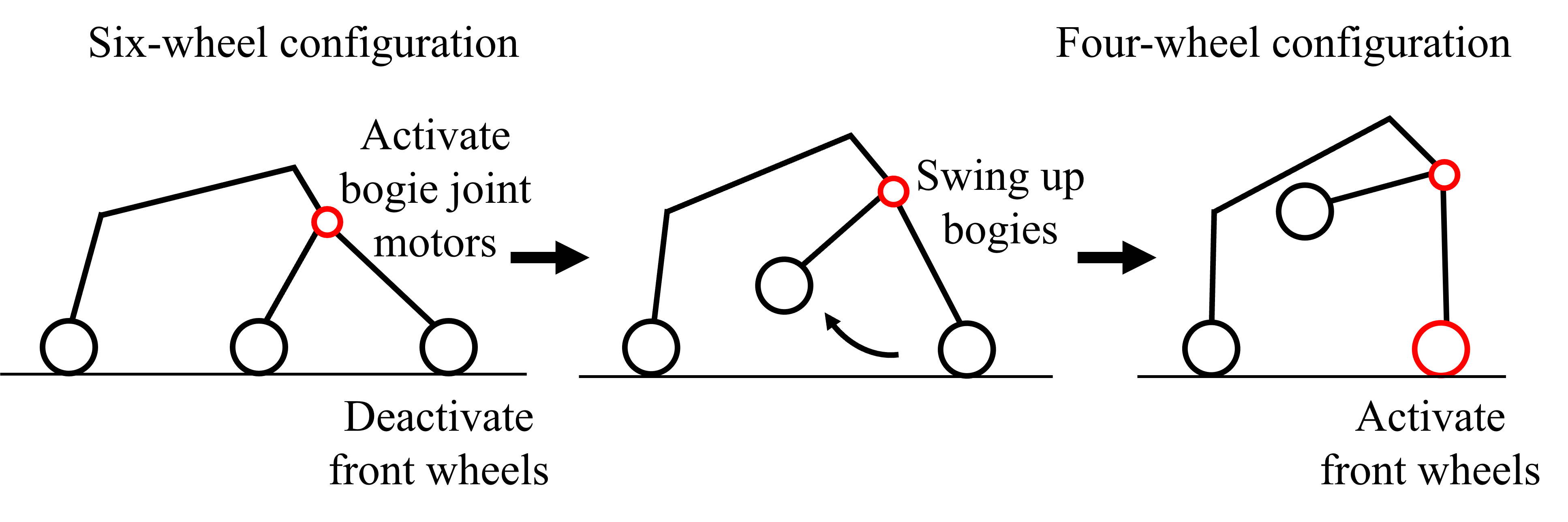}
        \subcaption{Six-wheel to four-wheel configuration transition\vspace{\baselineskip}}
    \end{minipage}
    \begin{minipage}{0.95\columnwidth}
        \centering
        \includegraphics[width=\columnwidth]{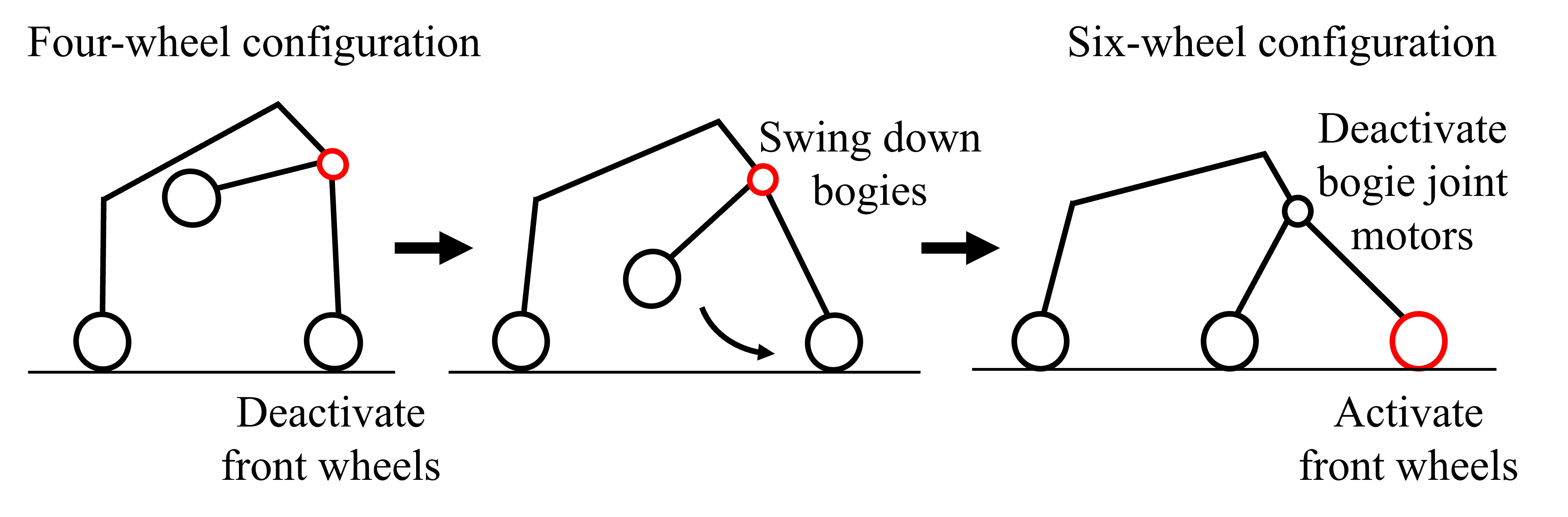}
        \subcaption{Four-wheel to six-wheel configuration transition}
    \end{minipage}
    \caption{Configuration change sequences}
    \label{fig:cs:confchange}
\end{figure}

Fig.~\ref{fig:cs:confchange} shows the motor control sequence executed when the robot performs reconfiguration.
When switching from the six-wheel configuration to the four-wheel configuration, the front wheels are first set to an underactuated state to reduce ground friction during the bogie swing-up motion. The bogie joint motors are then activated to raise the bogies, after which the control of the front wheels is reactivated.
Conversely, when transitioning from the four-wheel configuration to the six-wheel configuration, the front wheels are again made underactuated before the bogies are rotated downward. 
Once the bogies are fully lowered, the bogie joint motors are deactivated, and the control of front wheels is re-enabled.
\insfig{1.0\columnwidth}{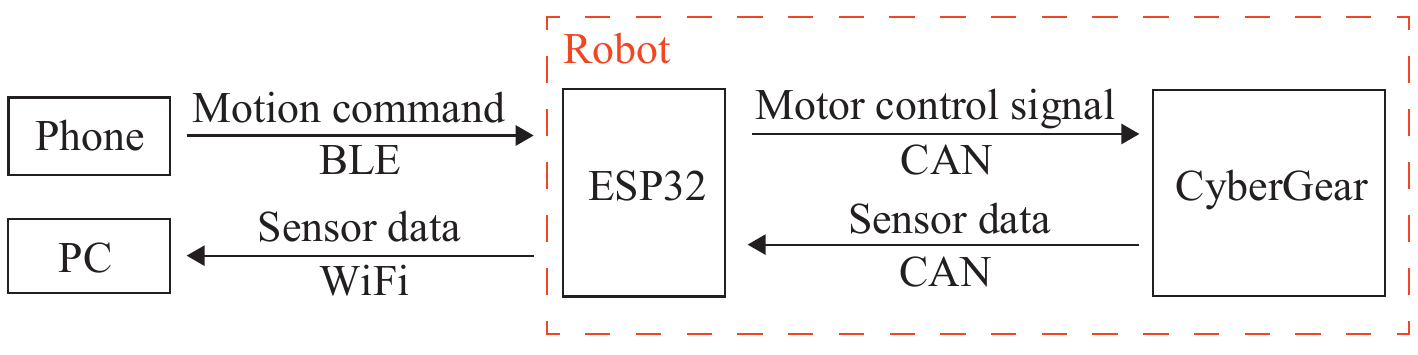}{Control system of the prototype robot}{fig:cs:system}

\section{EXPERIMENT}
To evaluate the turning performance and step-climbing capability of the proposed mechanism, we conduct experiments on turning motion and step climbing. 

\begin{figure*}[t]
    \centering
    \begin{minipage}{1.0\columnwidth}
        \centering
        \includegraphics[width=\linewidth]{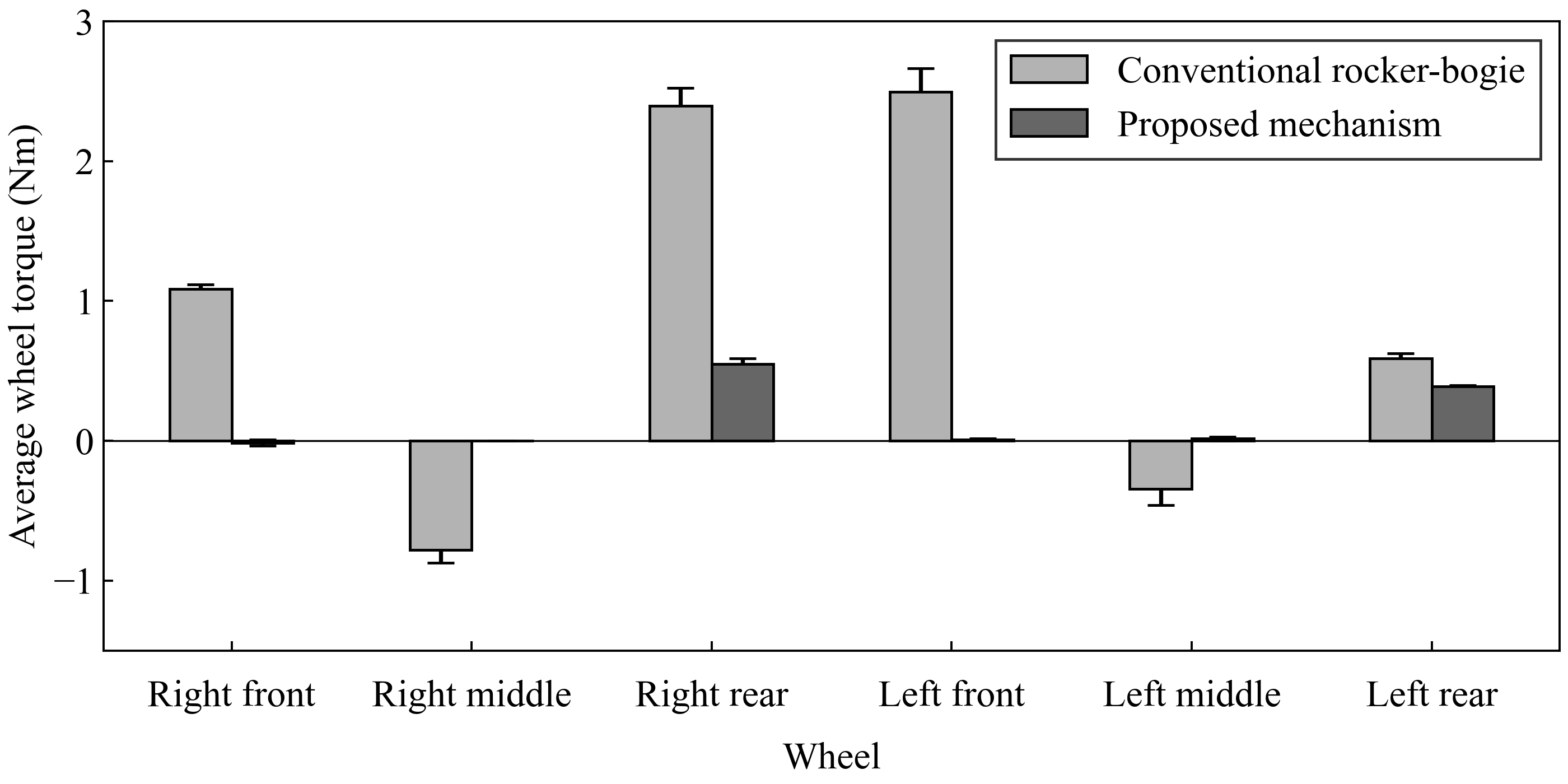}
        \subcaption{Counterclockwise rotation}
    \end{minipage}
    \hfill
    \begin{minipage}{1.0\columnwidth}
        \centering
        \includegraphics[width=\linewidth]{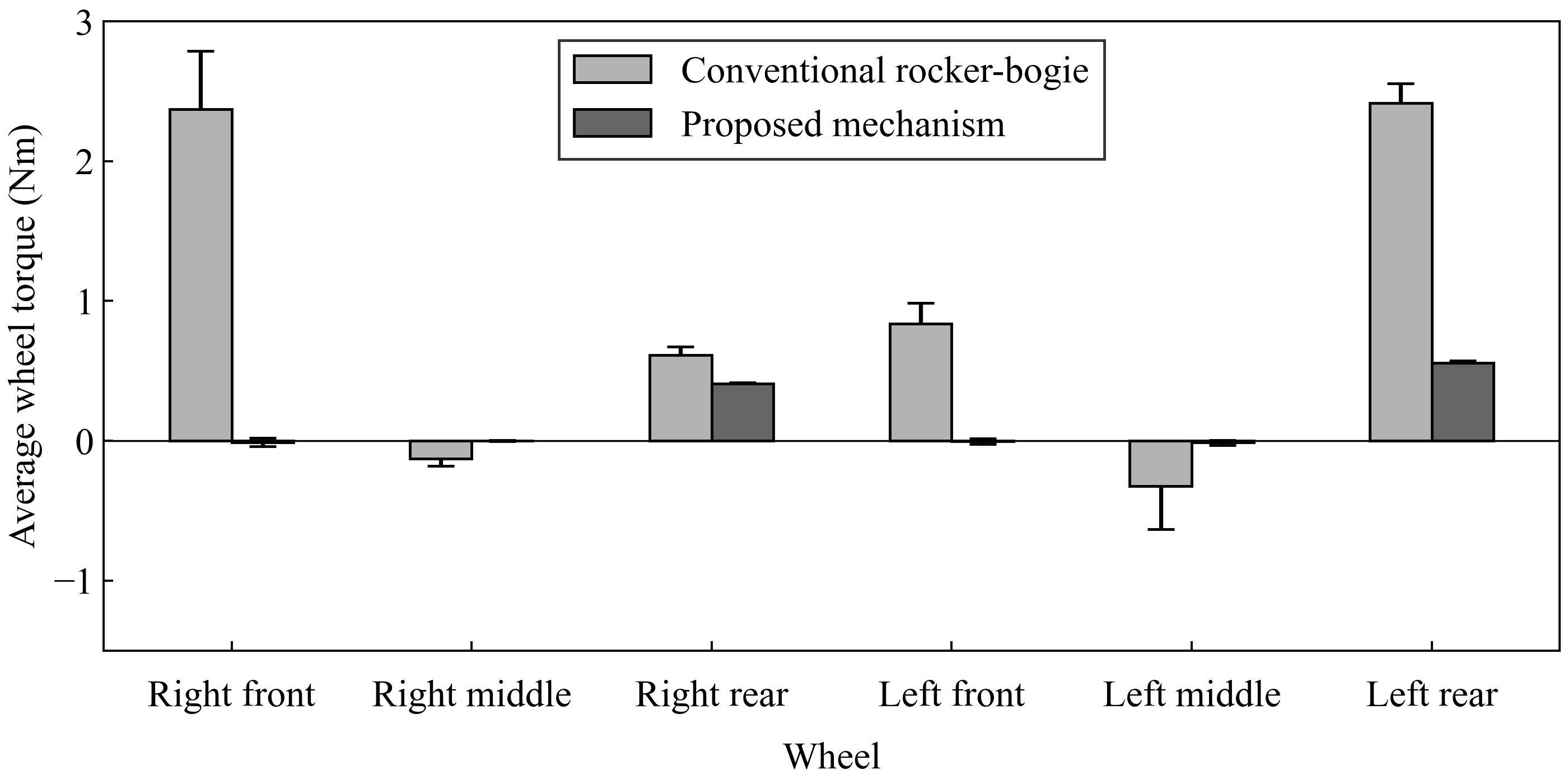}
        \subcaption{Clockwise rotation}
    \end{minipage}
    \vspace{-1mm}
    \caption{Comparison of wheel torques during turning. Positive values indicate torque contributing to the direction of rotation.}
    \vspace{-2mm}
    \label{fig:exp:trqcomp_r}
\end{figure*}

\subsection{Turning motion evaluation}
In the turning motion experiments, we apply a zero-radius turning velocity to the robot, as shown in Fig.~\ref{fig:cs:cmdsdr} (b). For comparison, the same experiments are also conducted using a conventional rocker-bogie mechanism equipped with six non-steerable grip wheels. We set the angular velocities of the front and rear wheels to 1, 3, and 5 rad/s, corresponding to three experimental conditions (r1, r3, and r5), respectively. For each condition, five clockwise and five counterclockwise turning trials are conducted for the proposed mechanism. In contrast, for the conventional rocker-bogie mechanism, only a single trial is performed because significant lateral slip resistance is expected to cause excessive wheel wear and impose high structural loads on the robot.
During the experiment, we record the motion using a top-view camera and determine the circumcircle radius and turning time from the recorded video. In addition, we measure the driving torque of each motor using the built-in sensing functionality of the CyberGear motors during turning. By comparing the results of both systems, we evaluate the advantages of the proposed mechanism. 

As a result, the circumcircle radii of the proposed mechanism and the conventional rocker-bogie mechanism with six non-steerable grip wheels were 77.7$\pm$0.6 cm and 72.8$\pm$2.2 cm, respectively. 
The times required for one complete rotation are shown in Table.~\ref{tab:One turning motion time}. 
Under the conventional rocker-bogie mechanism, the clockwise trials for r1 and r3 were terminated before completion because the bogie rotated after the robot began turning, causing the middle wheel to lift off the ground and altering the wheelbase between the front and rear wheels.
Abrasion of the wheel rubber was also observed.
Fig.~\ref{fig:exp:trqcomp_r} shows average wheel torque measured during all trials for each mechanism (conventional vs proposed). 
The left figure shows the average torque during counterclockwise turning, while the right figure presents the average torque during clockwise turning.
The conventional mechanism exhibits a maximum wheel torque exceeding 2 Nm, whereas the proposed mechanism shows approximately 0.5 Nm.
The total averaged wheel torque during counterclockwise turning was 5.44 Nm for the conventional mechanism and 0.95 Nm for the proposed mechanism. During clockwise turning, the corresponding values were 5.79 Nm and 0.94 Nm, respectively.

\subsection{Step-climbing evaluation}
In the step-climbing experiments, the prototype robot traverse a 40 cm step five times under the ``go straight'' command, as shown in Fig.~\ref{fig:cs:cmdsdr} (a).
The angular velocities of the middle wheels are set to 1, 3, and 5 rad/s, while those of the front and rear wheels are set to twice these values, corresponding to three experimental conditions (d1, d3, and d5), respectively. We define the initial state such that the front wheels of the robot are 30 cm apart from the step wall. The step-climbing time is determined as the duration from the initial state until the rear wheels reach the top surface of the step. We record the experiments and identify the step-climbing time and success rate from the recorded videos.

\begin{table}
    \centering
    \caption{One turning motion time}
    \vspace{-2mm}
    \begin{tabular}{c|ccc}
        \hline \hline
         & r1 & r3 & r5 \\
         \hline
         Conventional rocker-bogie & 165.28 & 50.27 & 34.58$\pm$1.00 \\
         Proposed mechanism & 24.70$\pm$0.33 & 8.61$\pm$0.14 & 5.24$\pm$0.12 \\
         \hline
    \end{tabular}
    \label{tab:One turning motion time}
\end{table}

\begin{figure*}
    \centering
    \vspace{1mm}
    \includegraphics[width=1\linewidth]{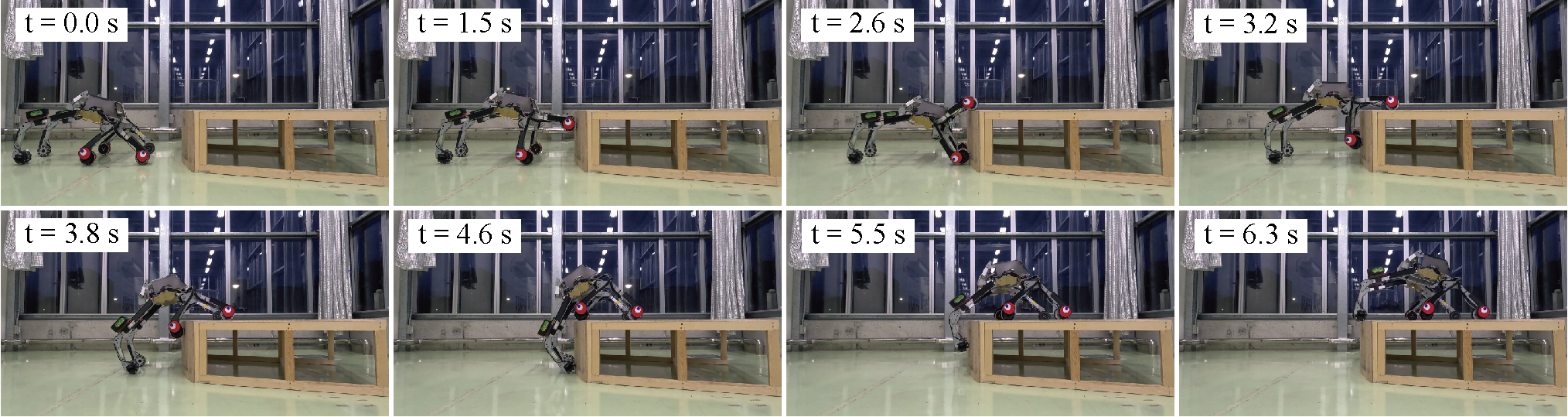}
    \caption{Step-climbing behavior of the prototype robot under condition d5}
    \label{fig:Step climbing}
    \vspace{-2mm}
\end{figure*}

Fig.~\ref{fig:Step climbing} shows the step-climbing behavior of the proposed mechanism. The success rates and required time of 40 cm step climbing are shown in Table.~\ref{tab:Result of the step climbing performance evaluation}. 
In the failed trials, the robot exhibited lateral tilting during step climbing, resulting in rollover.

\section{DISCUSSION}
From the turning motion experiments, the proposed mechanism enabled stable zero-radius turning. As shown in Table.~\ref{tab:One turning motion time}, it achieved turning speeds more than five times higher than those of the conventional rocker-bogie mechanism equipped with six non-steerable grip wheels. 
Furthermore, Fig.~\ref{fig:exp:trqcomp_r} indicates that the proposed mechanism requires substantially lower total average torque, corresponding to approximately 17\% of that of the conventional mechanism.
These results indicate that the proposed mechanism enables more efficient turning. 

Although the circumscribed turning radius of the conventional rocker-bogie mechanism was approximately 5 cm smaller, this reduction is attributed to lateral wheel slip shifting the instantaneous center of rotation.
In contrast, the proposed mechanism exhibits lower
variation in turning radius, indicating more consistent and geometrically predictable turning behavior.
This may be beneficial for reliable operation in practical navigation tasks.

Fig.~\ref{fig:exp:trqcomp_r} also reveals differences in torque distribution between the two mechanisms. The proposed mechanism exhibits higher motor torque at the rear wheels, whereas the conventional mechanism with six non-steerable grip wheels shows larger torque at diagonally opposite wheels. This suggests that torsional loads are induced in the structure of the conventional mechanism during turning, indicating that structural stiffness against torsional deformation should be carefully considered\cite{Shamah1999, Alamdari2014}. Additionally, the torque magnitude varies with turning direction. The prototype robot has an asymmetric structure: the left rocker is fixed to the chassis, while the right rocker is connected via a single pivot joint. This structural asymmetry likely affected force transmission from the wheels to the chassis, resulting in the bogie rotation observed in the clockwise r1 and r3 trials of the conventional mechanism.

As shown in Fig.~\ref{fig:Step climbing}, the step-climbing experiments confirmed that the proposed mechanism can climb a 40 cm step within a short time. 
As summarized in Table.~\ref{tab:Result of the step climbing performance evaluation}, higher wheel speeds reduce climbing time but also decrease success rate. At higher speeds, insufficient frictional engagement with the step wall causes slip, resulting in asynchronous climbing of the left and right front wheels. This misalignment alters the robot's orientation and leads to lateral tipping. 

Furthermore, we deployed the prototype robot in the XROBOCON robotics competition held at Expo 2025 Osaka. The competition field included 20 cm, 30 cm, and 40 cm steps, requiring rapid mobility and step-climbing capability. Excluding device-specific errors, the prototype robot consistently climbed all steps and utilized its high maneuverability to efficiently score points, ultimately winning the final competition\footnote{https://www.youtube.com/watch?v=s3FjQhGHlQs}. These results suggest that the proposed mechanism is particularly effective for environments requiring both step climbing and turning.

The proposed system introduces bogie joint motors to the rocker-bogie mechanism to change its configuration.
Following the concept proposed by Lim et al.\cite{Lim2022-sx}, the bogie joint motors could potentially be removed by mechanically transmitting the middle-wheel driving force to the bogie joint through a belt-driven or equivalent transmission system.
In addition, a locking mechanism could fix the bogie posture after swing-up, removing the need for holding torque in the four-wheel configuration and reducing energy consumption.

\begin{table}
    \centering
    \caption{Result of the step-climbing evaluation}
    \vspace{-2mm}
    \begin{tabular}{c|cc}
    \hline \hline
         speed & success rate & time (s)\\
         \hline
         d1 & 5/5 & 29.31 $\pm$ 0.12 \\
         d3 & 5/5 & 10.24 $\pm$ 0.31 \\
         d5 & 3/5 & 6.40 $\pm$ 0.12 \\
         \hline
    \end{tabular}
    \label{tab:Result of the step climbing performance evaluation}
\end{table}

\section{CONCLUSION}
This study proposed a reconfigurable rocker-bogie mechanism equipped with six wheel motors and two actuated bogie joints, achieving both high step-climbing capability and superior turning performance with a small number of actuators.
The proposed mechanism adaptively changes its configuration according to locomotion demands, enabling both six-wheel step climbing and four-wheel turning.
Experimental results demonstrated that the prototype robot achieves zero-radius turning at a speed more than five times that of a conventional rocker-bogie mechanism equipped with six non-steerable grip wheels, while requiring only approximately 17\% of the total average wheel torque.
The robot was also verified to climb a 40 cm step with an average climbing time of 6.4 s. 
These results indicate the potential of the proposed mechanism for mobile robotic applications requiring both terrain adaptability and agile maneuverability, such as logistics warehouses and industrial plants.

\section*{Acknowledgment}
The authors would like to express their sincere gratitude to all members of Team MarsMars and Beaver's Hive for their contributions to the development of the robot. The authors also gratefully acknowledge NHK Enterprises, Inc. for providing the opportunity to develop this robot, as well as all individuals involved in the planning and operation of the XROBOCON competition.
In the preparation of this letter, we used generative AI as a writing aid for language editing and formatting. All research content, results, and discussions were reviewed and verified by the authors, who take full responsibility for their accuracy.

\bibliographystyle{IEEEtran}
\bibliography{bibtex_xrobocon}
\vspace{12pt}

\end{document}